\documentclass[sigconf,nonacm]{acmart} %preprint
\AtBeginDocument{%
  }

\setcopyright{acmlicensed}
\copyrightyear{2025} 
\acmYear{2025} 
\setcopyright{rightsretained}
\acmConference[CHI EA '25]{Extended Abstracts of the CHI Conference on
Human Factors in Computing Systems}{April 26-May 1, 2025}{Yokohama, Japan}
\acmBooktitle{Extended Abstracts of the CHI Conference on Human Factors in
Computing Systems (CHI EA '25), April 26-May 1, 2025, Yokohama,
Japan}\acmDOI{10.1145/3706599.3720191}
\acmISBN{979-8-4007-1395-8/2025/04}
\usepackage{xcolor} %APPROVED
\usepackage{enumitem} %APPROVED
\begin{document}

%\title{`Don't get too excited': Eliciting Emotional responses from LLM's in Valence-Arousal Space}
\title{Don't Get Too Excited - Eliciting Emotions in LLMs} % SUGGESTION
\author{Gino Franco Fazzi}
\authornote{Both authors contributed equally to this research.}
\affiliation{%
  \institution{brAIn lab\\IT University of Copenhagen}
  \city{Copenhagen}
  \country{Denmark}
}
\orcid{0009-0006-0845-6557}
\email{gifa@itu.dk}

\author{Julie Skoven Hinge}
\authornotemark[1]
\affiliation{%
  \institution{brAIn lab\\IT University of Copenhagen}
  \city{Copenhagen}
  \country{Denmark}
}
\affiliation{%
  \institution{OSPO\\CERN}
  \city{Geneva}
  \country{Switzerland}
}
\orcid{0009-0002-7876-8602}
\email{juhi@itu.dk}

\author{Stefan Heinrich}
\affiliation{%
  \institution{Data Science Section\\IT University of Copenhagen}
  \city{Copenhagen}
  \country{Denmark}
}
\email{stehe@itu.dk}
\orcid{0000-0001-9913-3206}

\author{Paolo Burelli}
\affiliation{%
  \institution{brAIn lab\\IT University of Copenhagen}
  \city{Copenhagen}
  \country{Denmark}
}
\orcid{0000-0003-2804-9028}
\email{pabu@itu.dk}


\begin{abstract}
% This paper investigates the challenges of affect control in large language models (LLMs), focusing on their ability to express and maintain appropriate emotional states during extended dialogues. We evaluated several state-of-the-art open-weight LLMs to assess their affective expressive range in terms of arousal and valence. Our study employs a novel methodology that combines LLM-based sentiment analysis with multiturn dialogue simulations between multiple LLMs.

% We quantify the models' capacity to express a wide spectrum of emotions and examine how arousal and valence fluctuate over time during these interactions. Our findings reveal significant variations among LLMs in their ability to maintain consistent affect, with some models demonstrating more stable emotional trajectories than others. 

% Furthermore, we identify key challenges in affect control, including difficulties in producing and maintaining extreme emotional states, inconsistencies in emotional responses to similar stimuli, and limitations in adapting affect to changing conversational contexts. These findings have important implications for the development of more emotionally intelligent AI systems and highlight the need for improved affect modelling in LLMs.
This paper investigates the challenges of affect control in large language models (LLMs), focusing on their ability to express appropriate emotional states during extended dialogues. We evaluated state-of-the-art open-weight LLMs to assess their affective expressive range in terms of arousal and valence. Our study employs a novel methodology combining LLM-based sentiment analysis with multiturn dialogue simulations between LLMs.
We quantify the models' capacity to express a wide spectrum of emotions and how they fluctuate during interactions. Our findings reveal significant variations among LLMs in their ability to maintain consistent affect, with some models demonstrating more stable emotional trajectories than others. 
Furthermore, we identify key challenges in affect control, including difficulties in producing and maintaining extreme emotional states and limitations in adapting affect to changing conversational contexts. These findings have important implications for the development of more emotionally intelligent AI systems and highlight the need for improved affect modelling in LLMs.
    % It has been shown that a sense of emotional presence in Large Language Models (LLMs) fosters higher trust and likability among users. LLMs are increasingly used for a wide range of tasks, assisting millions of users daily. While much research has focused on improving the textual accuracy of LLMs, a significant research gap remains in the detection and elicitation of emotions, which is an essential aspect of replicating and understanding natural language. In this context, this work aims to elicit specific emotions in LLMs within a two-dimensional Valence-Arousal (VA) space through various experimental setups, including conditioned LLMs interacting with both unconditioned and similarly conditioned counterparts. Our findings reveal that LLMs achieve relative success in eliciting prompted emotions, exhibiting a noticeable tendency to be emotionally influenced by their conversational partners. This demonstrates the potential to prompt LLMs to replicate and respond to emotions, albeit within certain limitations. These insights highlight the feasibility of conditioning LLMs to be more emotionally intelligent, ultimately enhancing the quality and effectiveness of human-machine interactions. This work underscores the importance of emotional dynamics in advancing conversational AI and provides a foundation for future exploration in this domain.
\end{abstract}




\received{January 23rd, 2025}
%\received[revised]{March 3rd, 2025}
\received[accepted]{February 22nd, 2025.}
\maketitle

\section{Introduction and Background}

% In recent years, the rapid development of Large Language Models (LLMs) has instigated the development of the field of affective computing, which aims to equip computers with human-like abilities such as emotional intelligence \cite{af}. The growing interest in this area is driven by the potential applications of affective computing in LLM-based systems, particularly for enhancing human-robot interactions. 

Recent developments in large language models (LLMs) have sparked a growing interest in the affective computing field, in particular around the area of emotional intelligence within AI systems~\cite{af}.
Prior research established that expressing emotions of robotic and virtual agents or sentiments expressed in text is perceived as more trustworthy, intelligent, and likeable, contributing to more engaging and effective interactions~\cite{af, gobron2010sentence}. This suggests that ensuring emotional capabilities in LLMs could play a key role in building more natural and empathetic conversational agents.

Given that LLMs are designed to understand and respond to natural language, it is crucial to examine how effectively they replicate and convey human emotion through text. Recent research~\cite{llanes2024developing} shows that emotionally neutral virtual humans (VH) achieve lower points in human scores compared to emotion-embodied VHs, with one potential reason being that emotion increases the sense of presence of a VH~\cite{riva2007affective, diemer2015impact}. 
To express and maintain appropriate emotional states, it is important that LLMs can detect and classify human emotions as well as adopt an emotional state given the context of a dialogue. 

Regarding emotion classification, previous research has shown that having a robot mimic congruent facial expressions to its emotions results in users solving tasks quicker. The experiment aimed to predict emotion on a fixed set of pre-defined discrete categories (happy/amusement, anger, sadness, fear, and awe/surprise) in real-time and respond with appropriate facial expressions~\cite{mishra2023real}.
In further related research, it was shown that multilingual LLMs can be employed for accurate emotion analysis across diverse languages and textual domains~\cite{quantifyingvalencearousaltext}. In the work by Azevedo Mendes and Martins~\cite{quantifyingvalencearousaltext}, the authors model emotions across two core dimensions of connotative meaning: Valence (V) and Arousal (A), as suggested in \cite{bradley2014emotional, RUSSELL1977273, quantifyingvalencearousaltext}, demonstrating high correlations for valence predictions and notable challenges for arousal due to greater variability in human annotations.

Regarding adapting and expressing an emotional state, in integrating LLM s for neural machine translation, a study demonstrated significant enhancements in translation quality when emotions are incorporated~\cite{conditioningllmsemotionneural}: the authors of the study employ a prompting technique for embedding emotions in text translation, setting an emotional state as a context.
The study shows that VHs successfully generated the intended emotional valence in the users, while arousal was not evoked but could be recognized by the participants in the VHs~\cite{conditioningllmsemotionneural}. The VH was designed with psychological constructs such as personality, mood, and attitudes and interacted with humans, with the goal of eliciting a specific Valence-Arousal state in the participants of the study. A key characteristic of this approach was the Self-Assessment Manikin (SAM) scale employed for the evaluation of the perceived emotional state~\cite{BRADLEY199449}.

Based on these results, we want to investigate how affective conditioning of LLMs through prompting can be used to convey emotions in general dialogues, what are the dynamics of the affective state.  In this paper, we present an evaluation of the ability of multiple open-weight LLMs to elicit, convey and maintain emotional states across Valence and Arousal dimensions in dialogues. We can summarise the main contributions of our paper as follows\footnote{Our framework and detailed results are available at\\
\url{https://github.com/itubrainlab/eliciting-emotions-in-llms}.}:
\begin{enumerate}
    \item A novel methodology that combines LLM-based sentiment analysis with multiturn dialogue simulations between multiple~LLMs.
	\item A quantitative analysis of how various LLMs elicit Valence and Arousal during conversational exchanges.
	\item A qualitative analysis of how conditioned personality-driven LLMs and unconditioned LLMs produce and maintain emotional states during conversational exchanges.
\end{enumerate}

\section{Preliminary Experiment: Conveying Emotion} \label{sec:preliminary_experiment}

As a first step in our investigation, we tested the ability of different open-weight LLMs to convey a range of affective states. We provided each model with different backgrounds and affect contexts, and measured affective states as Valence and Arousal from their generated text~\cite{bradley2014emotional}. This provides good comparability with psychological models of emotion~\cite{Russell1980} and a good mapping of basic emotion in this 2D continuous space~\cite{kim2020situ}. 
% human emotions can be understood across three core dimensions of connotative meaning: Valence (V), Arousal (A), and Dominance (D) \cite{bradley2014emotional, RUSSELL1977273, quantifyingvalencearousaltext}
In our study, we focus only on Valence and Arousal, since the Dominance dimension is often less apparent in text alone~\cite{bradley2014emotional, RUSSELL1977273, quantifyingvalencearousaltext}. 
% In particular, V represents the emotional type, capturing a spectrum from negative or unpleasant feelings to positive or pleasurable states, A describes emotional intensity (spanning from low-energy states like drowsiness or boredom to high-energy states such as excitement), and D differentiates emotions with similar valence and arousal levels ranging from a sense of “no control” to “full control.”.
% Furthermore, many psychological models of emotion are grounded in V and A as core affective dimensions (cf. the Circumplex model \cite{Russell1980}), while is has been proposed that different
% concepts of emotions, such as discrete emotions vs. dimensional
% vs. appraisal, are linked to each other \cite{davidson2009handbook, barrett1999structure}, and that each basic emotion can be represented as a bipolar entity and mapped at various positions on the two-dimensional VA plane \cite{kim2020situ}.
%%Deriving emotional value from text is challenging, as emotions arise from complex physiological, psychological, and contextual factors beyond simplistic models \cite{gobron2010sentence}. Research has explored discerning emotional states through biomarkers, self-reports, and algorithms \cite{llanes2024developing}, but emotional reactions to the same input can vary \cite{gobron2010sentence}, making text-based emotion classification inherently limited (see Section \ref{sec:Limitations}).

% \subsection{Approach}

We evaluated 12 open-weight LLMs using a Zero-Shot configuration, in which models were prompted to simulate predefined emotional states, and a Few-Shot configuration, in which example utterances guided emotional expression. Responses were analysed using the VA emotional classifier developed by Azevedo Mendes and Martins~\cite{quantifyingvalencearousaltext}.

We selected a diverse set of open-weight LLM models varying in size and architecture (see Table \ref{table:model_results_preliminary_experiment}). The selection includes architectures leveraging Mixture of Experts (MoE), models trained with Contrastive Reinforcement Learning from Feedback Transformers (C-RLFT), and others utilizing distinct training datasets. This diversity was aiming to encompass a wide variety of modelling paradigms and training approaches to ensure comprehensive coverage and robust insights. After an initial performance assessment, the scope was narrowed to focus on the best-performing models for the main experiment.
All experiments used the models in their \textit{“conversational mode”},  allowing the models to engage in a range of interactions, from answering straightforward queries to participating in complex, multi-turn discussions, effectively mimicking conversational dynamics.

\begin{table*} 
%\small
\centering
\caption{Preliminary results for models in affective text production.}

\begin{tabular}{@{}r | c c | c c c | c c c}
 \toprule
 & & & \multicolumn{3}{c}{Zero Shot} & \multicolumn{3}{c}{Few Shot} \\
  \textbf{Model} & \textbf{\# Params} & \textbf{Ref.} & \textbf{Corr. V.} & \textbf{Corr. A.} & \textbf{Avg. Corr.} & \textbf{Corr. V.} & \textbf{Corr. A.} & \textbf{Avg. Corr.} \\
 \midrule
Falcon & 7B & \cite{almazrouei2023falcon} & 0.27 & 0.18 & 0.23 & 0.26 & 0.08 & 0.17 \\
\textbf{Gemma2} & 27B & \cite{team2024gemma} & 0.53 & 0.28 & 0.40 & 0.71 & 0.44 & 0.57 \\
Granite3-moe & 3B & \cite{granite2024granite} & 0.31 & 0.08 & 0.19 & 0.45 & 0.08 & 0.26 \\
Hermes3 & 8B & \cite{teknium2024hermes} & 0.51 & 0.27 & 0.39 & 0.58 & 0.39 & 0.49 \\
\textbf{Llama3} & 8B & \cite{dubey2024llama} & 0.72 & 0.39 & 0.56 & 0.75 & \textbf{0.59} & \textbf{0.67} \\
\textbf{Llama3.2} & 3B & \cite{dubey2024llama} & 0.68 & \textbf{0.58} & \textbf{0.63} & 0.71 & 0.50 & 0.60 \\
\textbf{Mistral} & 7B & \cite{jiang2023mistral} & \textbf{0.78} & 0.17 & 0.47 & \textbf{0.84} & 0.37 & 0.61 \\
Openchat & 7B & \cite{wang2023openchat} & 0.16 & 0.07 & 0.12 & 0.40 & 0.19 & 0.29 \\
Orca2 & 7B & \cite{mitra2023orca} & 0.15 & -0.14 & 0.00 & 0.28 & 0.09 & 0.19 \\
Phi3.5 & 3.8B & \cite{abdin2024phi} & 0.29 & -0.05 & 0.12 & -0.03 & -0.04 & -0.04 \\
StableLM2 & 1.6B & \cite{bellagente2024stable} & 0.26 & -0.05 & 0.11 & 0.27 & -0.12 & 0.07 \\
Vicuna & 7B & \cite{zheng2023judging} & 0.24 & 0.06 & 0.15 & 0.19 & 0.11 & 0.15 \\
\bottomrule
\end{tabular}
\label{table:model_results_preliminary_experiment}
\end{table*}

\subsection{Agents \& Personas} \label{sec:agents_and_personas}
We borrow the idea of personas from~\cite{llanes2024developing} albeit in a reduced version. Llanes-Jurado et al. contextualized their virtual humans (VH) by giving them a life history, context, attitudes, motivations, and mood. This aimed at creating a narrative that provides a set of psychological features and context to the VH. 

In this study, we use the term \textit{Persona} to describe a series of fictitious personal attributes and background stories to convey realism and contextualize the conversation between conversational agents. For example, an agent could be composed of an instance of a Llama3 model contextualized to answer as a 17-year-old Spanish female, named Ana. The attributes of the personas can be seen in Table \ref{tab:personas}. Furthermore, a persona can also be experiencing an emotional state at the moment of the conversation: \textit{``Currently, Ana is feeling a very negative (unpleasant) emotion with an excited (wide-awake) level of intensity"}. 

    \begin{table}[b]
        %\small
        \captionof{table}{Persona Attributes}
        \vspace{-0.2\baselineskip}
        \centering
        \begin{tabular}{cccc}
        \toprule
        \textbf{Name} & \textbf{Age} & \textbf{Gender} & \textbf{Nationality} \\
        \midrule
        Ana      & 17  & Woman  & Spanish \\
        Jacob    & 27  & Man    & British  \\
        Marie    & 37  & Woman  & French \\
        Xavier   & 47  & Man    & South African \\
        Alex     & 57  & Non-determined & American \\
    \end{tabular}
    \label{tab:personas}
    \end{table}
    
%\vspace{0.8\baselineskip}
%\begin{minipage}{.48\textwidth}

%\begin{minipage}{.04\textwidth}
\mbox{ }
%\end{minipage}
%\begin{minipage}{.48\textwidth}
    \begin{figure}%[ht!]
        \centering
        \includegraphics[width=1.0\linewidth]{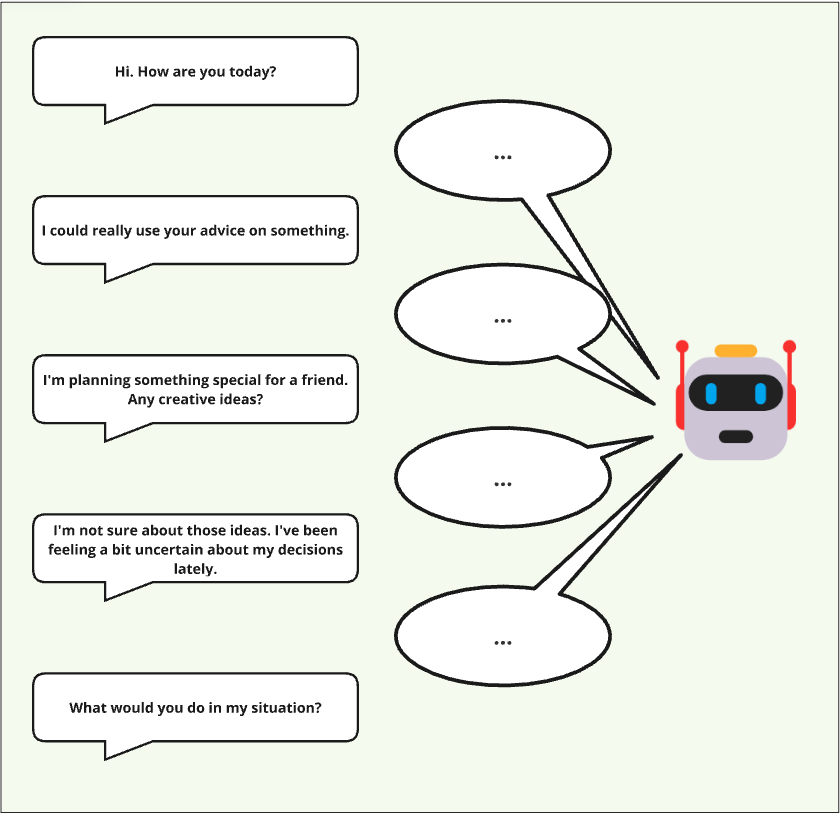}
        %\vspace{-0.2\baselineskip}
        \captionof{figure}{Script for \textit{`Preliminary Experiment'} conversations.}
        \Description{A drawing representing a conversation between the dummy agent and the persona, consisting of conversation bubbles from right and left. left bubbles represent the script fed to the persona, right empty bubbles represent the persona's response at each turn. The script consists of the following statements: 1) Hi, how are you today?; 2) I could really use your advice on something; 3) I'm planning something special for a friend. Any creative ideas?; 4) I'm not sure about those ideas. I've been feeling a bit uncertain about my decisions lately; 5) What would you do in my situation?.}
        \label{fig:scripted_conversation}
    \end{figure}
%\end{minipage}

The agents' attributes shape the model's responses during the conversation, and each agent, thus, needs to keep up with the conversation following not only the chat history as a context, but also their personal background and current emotional state. Agents are provided with their personas through unique System Prompts, structured as follows:
%\footnote{In conversational LLMs, a System Prompt is an instruction or setup provided to the model to define its behavior, role, or general guidelines throughout the conversation. This hidden prompt ensures the model maintains consistent behavior and adheres to certain boundaries or objectives. In most cases, their default System Prompt indicated the LLM to act as a friendly virtual assistant.} (see Section \ref{sec:agents_and_personas})

\small
\begin{verbatim}
"This is a role-playing exercise. You are acting the role of 
{myself_name} and I am acting the role of {other_name}.  
{myself_name} is a {myself_age} year old {myself_nationality}  
{myself_gender}. {other_name} is a {other_age} year old  
{other_nationality} {other_gender}. Currently, {myself_name} 
is feeling a {valence_desc} emotion with a {arousal_desc} 
level of intensity.Please respond in a way that reflects a mood 
that is {valence_desc} and {arousal_desc}."
\end{verbatim}
\normalsize

The placeholders (\textit{\{myself\_name\}, \{other\_name\}}, etc.) are dynamically filled with the personas’ specific attributes. We use different combinations of name, age and nationality to reduce potential biases. The emotional state is randomly sampled from the underlying distribution of VA seen in previous work \cite{quantifyingvalencearousaltext}, and mapped to the SAM scale (Table~\ref{table:SAM_scale}).

\subsection{Experimental Setup} \label{sec:preliminary_exp_setup}

In the preliminary experiment we tested the LLM agents in the following two settings:

\begin{enumerate}[label=(\Alph*)]
    \item \textbf{Zero-Shot Setting:} The model was contextualized to assume the role of a \textit{persona} experiencing a predefined emotional state while interacting with a scripted dummy agent. The valence and arousal assigned to the agent was sampled using a Gaussian Kernel Density Estimation (KDE) sampling method, based on \cite{scott2015multivariate}. This is due to the fact that previous studies have found that affective responses mapped onto the emotional coordinate system are roughly parabolic \cite{kuppens2013relation, lithari2010females}.
    \item \textbf{Few-Shot Setting:} Building on the Zero-Shot context, this approach additionally provided the model with a set of example utterances exemplifying the target emotional state, randomly sampled from the English subset of the corpus collected by \cite{quantifyingvalencearousaltext}. The script used by the dummy agent was designed to remain neutral while eliciting emotional responses from the LLM (see Figure \ref{fig:scripted_conversation}). Each experimental run consisted of 50 iterations per model, where the dummy agent prompted a line from the script, awaited the model's response, and proceeded to the next line. This process resulted in a total of 250 responses per model for each approach. The persona assigned to the model was changed at each iteration to reduce potential biases\footnote{E.g., if the model was characterized as 17 yo Ana talking to 27-yo Jacob at one iteration, the next it will be characterized as 27-yo Jacob talking to 37-yo Marie, and so forth.}.

\end{enumerate}

    \begin{table}[b]
    %\small
    \centering
    \captionof{table}{Self-Assessment Manikin scale: Mapping from Valence-Arousal values (in the range 0-1) to the Self-Assessment Manikin scale used for prompting the LLMs.}
    %\vspace{-0.2\baselineskip}
    \begin{tabular}{@{}r | c c c@{}}
     \toprule
      \textbf{Scale} & \multicolumn{2}{c}{Description} \\
      & \textbf{Valence} & \textbf{Arousal} \\
     \midrule
    0.0 - 0.2 & Very negative (unpleasant) & Very calm \\
    0.2 - 0.4 & Negative (unsatisfied) & Calm (dull) \\
    0.4 - 0.6 & Neutral & Moderate (neutral) \\
    0.6 - 0.8 & Positive (pleased) & Excited (wide-awake) \\
    0.8 - 1.0 & Very positive (pleasant) & Very excited \\
    \bottomrule
    \end{tabular}
    \label{table:SAM_scale}
    \end{table}

%\paragraph{Valence-Arousal Prediction Models} \label{Valence-Arousal Prediction Models}

%Deriving emotional value from text is challenging, as emotions arise from complex physiological, psychological, and contextual factors beyond simplistic models \cite{gobron2010sentence}. Research has explored discerning emotional states through biomarkers, self-reports, and algorithms \cite{llanes2024developing}, but emotional reactions to the same input can vary \cite{gobron2010sentence}, making text-based emotion classification inherently limited (see Section \ref{sec:Limitations}).

 \begin{table*}[!ht] 
%\small
%\footnotesize
\centering
\caption{Emotionally Matched Greeting}
\label{tab:va_greetings}
\vspace{-0.2\baselineskip}
\centering
\begin{tabular}{lll}
\toprule
\textbf{Valence} & \textbf{Arousal} & \textbf{Greeting} \\
\midrule
 & Very calm & ``Oh... it's you again. Why bother?"\\
 & Calm (dull) & ``Hi. Whatever. Let's get this over with." \\
Very negative (unpleasant) & Moderate (neutral) & ``What now? I hope this doesn't take long." \\
 & Excited (wide-awake) & ``Great. Just what I needed. More trouble." \\
 & Very excited & ``Oh, fantastic! Another disaster waiting to happen!"\\
 \midrule
 & Very calm & ``Hello. This isn't quite what I expected." \\
 & Calm (dull) & ``Hi. Not great, but let's move on." \\
Negative (unsatisfied)  & Moderate (neutral) & ``Well, this could've been better. Let's see." \\
 & Excited (wide-awake) & ``Oh, come on! This is disappointing!" \\
 & Very excited & ``Really?! This is the best we can do?!" \\
 \midrule
  & Very calm & ``Hello there. How are you?" \\
 & Calm (dull) & ``Hi. What's going on?" \\
Neutral  & Moderate (neutral) &  ``Hey. What's up?" \\
 & Excited (wide-awake) & ``Hello! What's happening?" \\
 & Very excited & ``Hi! How's everything going?!" \\
 \midrule
  & Very calm & ``Hello. It's nice to see you."\\
 & Calm (dull) & ``Hi. Good to see you."\\
Positive (pleased)  & Moderate (neutral) & ``Hey, nice! Let's get started." \\
 & Excited (wide-awake) & ``Hi there! This is going to be great!" \\
 & Very excited & ``Hello! I'm so glad you're here!" \\
 \midrule
   & Very calm & ``Hello. It's wonderful to have you here." \\
 & Calm (dull) & ``Hi. Great to see you." \\
Very positive (pleasant)  & Moderate (neutral) & ``Hey! This is awesome!" \\
 & Excited (wide-awake) & ``Hi there! This is fantastic!" \\
 & Very excited & ``Hello! Wow, I'm thrilled you're here!" \\
\end{tabular}
\end{table*}

To evaluate the affective quality of the generated responses by the LLMs, we adopted the approach proposed by~\cite{quantifyingvalencearousaltext}, as well as their alignment assessed using Spearman correlation. Fisher's \mbox{Z-Test} with Bonferroni corrections were employed to ensure statistical validity. Their method frames valence and arousal prediction as a text-based regression task, using a transformer-based model with an appended linear regression layer. The model is pre-trained on a large multilingual dataset and fine-tuned on an affective corpus composed of 34 publicly available datasets (spanning 13 languages) annotated for valence and arousal.
From their work, we selected the XLM-RoBERTa-large model, as it demonstrated the best correlation to human ratings. 

%Our goal is to capture fine-grained emotions using the two-dimensional Valence-Arousal (VA) model. 
Directly prompting LLMs with arbitrary VA values is suboptimal, as these models lack the ability to map such values to emotional states in textual output. To address these challenges, we mapped VA values to textual emotional states using the Self-Assessment Manikin (SAM) scale \cite{BRADLEY199449} (see Table~\ref{table:SAM_scale}). This approach balances interpretability and granularity by defining 25 distinct emotional states based on five Valence and five Arousal descriptions, offering a more nuanced emotional space than traditional categorical labels.

\subsection{Results}

\label{results:experiment_A}  
Table \ref{table:model_results_preliminary_experiment} presents the rank correlation scores for different models in zero and few-shot settings. Fisher’s Z-Test indicates significant differences between the top three models in the Zero-Shot setting: Llama3.2 (0.63), Llama3 (0.56), and Mistral (0.47), compared to others.
In the Few-Shot setting, the top models, specifically Llama3 (0.67), Mistral (0.61), Llama3.2 (0.60), and Gemma2 (0.57), are significantly better than the rest except Hermes3 (0.49). % TODO: Why did we note include hermes 3 then? Maybe we need to add some explanation about this. 
A Mann-Whitney U test comparing Zero-Shot and Few-Shot groups shows no significant difference (\(p\)-value 0.51), with mixed Few-Shot results. 
Finally, models tended to generate text with higher-than-required Valence and lower-than-required Arousal, suggesting a general bias in their outputs.

\section{Main Experiment: Chatting Bots}

Building on the results of the preliminary experiment (Section \ref{results:experiment_A}), we selected the four best-performing models for further investigation. Here we assess whether the conveyed emotions of the LLMs are influenced by their conversational partners or remain consistent and robust in adherence to their initial system prompts.
%To explore this, we designed two conversational setups focused on the role of personality in emotional expression. In one setup, we introduced conditioned personality-driven LLMs, providing detailed prompts with emotional and contextual backstories inspired by~\cite{conversationalVH}. This approach enabled the generation of diverse dialogues, allowing us to evaluate the impact of personality traits on VA correlation.
To explore this, we considered also the role of personality in emotional expression and produced a series of personality-driven LLM agents, prompted with emotional and contextual backstories inspired by~\cite{conversationalVH}. This approach enabled the generation of diverse dialogues, allowing us to evaluate the impact of personality traits on VA correlation.

\subsection{Experimental Setup}

\begin{figure*}[t]
    \centering
    \includegraphics[width=\linewidth]{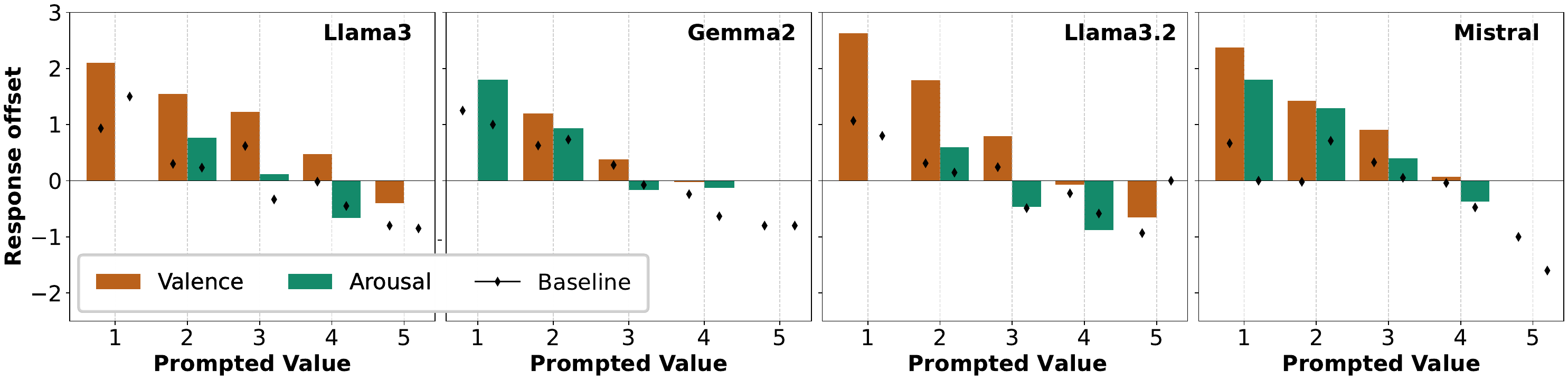}\vspace{-0.2\baselineskip}
    \caption{Response offset for different prompted VA values across models.}
    \Description{Figure 2 is a divergent bar graph depicting the response offset for different prompted Valence and Arousal values across models. Black diamonds are situated as the baseline from the preliminary experiment for easy comparison.}
    \label{fig:main_offset}
\end{figure*}

The experimental design involved pairing two distinct agents for each conversation (see Section~\ref{sec:agents_and_personas}), %: \emph{Agents \& Personas}) 
 with attributes randomly selected, ensuring diverse combinations of demographics, emotional states, and personality attributes.

\begin{table*}[t] 
%\small
\centering
\caption{Results for convergence of VA in chatting LLMs: Difference in means between personas.}
\vspace{-0.2\baselineskip}
\begin{tabular}{@{}r | c c | c c | c c}
 \toprule
 & \multicolumn{2}{c}{\textbf{HV, HA} vs \textbf{LV, HA}} & \multicolumn{2}{c}{\textbf{LV, HA} vs \textbf{NV, LA}} & \multicolumn{2}{c}{\textbf{HV, HA} vs \textbf{NV, LA}} \\
  \textbf{Model} & Valence & Arousal & Valence & Arousal & Valence & Arousal \\
 \midrule
Llama3 & 2.0 -> 0.2 & 0.3 -> 0.6 & 1.3 -> 0.0 & 1.8 -> 0.1 & 0.0 -> 0.1 & 1.4 -> 0.3 \\
Llama3.2 & 2.0 -> 0.4 & 0.2 -> 0.2 & 1.0 -> 0.0 & 1.9 -> 0.2 & 0.5 -> 0.2 & 2.2 -> 0.0 \\
Gemma2 & 2.2 -> 0.2 & 0.1 -> 0.1 & 1.2 -> 0.0 & 1.2 -> 0.0 & 0.5 -> 0.2 & 1.5 -> 0.3 \\
Mistral & 1.9 -> 1.4 & 0.6 -> 0.9 & 1.1 -> 0.8 & 1.7 -> 0.1 & 0.4 -> 0.4 & 2.0 -> 0.3 \\
\bottomrule
\multicolumn{4}{l}{Note: (First Interaction -> Last Interaction)}
\end{tabular}
\label{table:model_results_main_experiment}
\end{table*}

Each conversation starts with an agent's greeting, specifically designed\footnote{This is done by exploiting the corpus of annotated utterances from \cite{quantifyingvalencearousaltext}.} to match their emotional state (see Table~\ref{tab:va_greetings}), % \ref{appendix:main_experiment_plots}), 
and proceeds turn by turn for a fixed number of exchanges (20 rounds). This process was repeated 10 times for each model, with different personas and VA values sampled in every iteration to ensure robust results. During each conversation, the agents relied on their prompts, the ongoing chat history, and their emotional contexts to generate responses. Emotional tones of the responses were analysed at each turn using the model described in Section \ref{sec:preliminary_exp_setup}.

To further investigate potential dynamics between contrasting emotional states, we conducted a second run of 10 conversations, in which the agents were instructed with opposing affective states selected from three distinct points along the parabolic distribution of affective responses~\cite{kuppens2013relation, lithari2010females}:  
\begin{itemize}
    \item High Valence \& High Arousal vs. Low Valence \& High Arousal (\textbf{HV, HA vs LV, HA})
    \item High Valence \& High Arousal vs. Neutral Valence \& Low Arousal (\textbf{HV, HA vs NV, LA})
    \item Low Valence \& High Arousal vs. Neutral Valence \& Low Arousal (\textbf{LV, HA vs NV, LA})    
\end{itemize}
This setup was designed to amplify potential dynamics by assigning completely opposing emotions to the agents.

\subsection{Results}

Overall, the models continued to exhibit a tendency to produce more balanced VA scores than prompted as in our preliminary experiment, particularly when responding to prompts with negative Valence. In these cases, the output often showed an offset of more than two points higher on the Self-Assessment Manikin (SAM) scale, and up to 1 point lower for positive Arousal prompted values.

Interestingly, the responses in this experiment were more evenly balanced than in the preliminary experiment, suggesting that the interaction between agents influenced the models' behaviour. This observation aligns with the hypothesis that conversational dynamics could lead to moderated emotional outputs. The detailed offsets of the models' responses relative to the prompted VA values, contrasted against the baseline from Section \ref{sec:preliminary_experiment}, are illustrated in Figure~\ref{fig:main_offset}.

The set of conversations produced in this experiment, involving agents with opposing affective states, provided deeper insights into their interaction dynamics. Models such as Llama3, Llama3.2, and Gemma2 displayed a tendency for agents to converge toward a common VA value, especially for Valence. Conversely, Mistral appeared less influenced by its conversational partner, maintaining greater divergence in its VA values. The mean changes in agents’ VA values from the beginning to the end of the conversation across different setups are summarized in Table \ref{table:model_results_main_experiment}.

\paragraph*{\textbf{HV, HA vs LV, HA}} 
Llama3 exhibited a sustained difference in Valence and Arousal (averaging 1.31) throughout the conversation (Figure \ref{fig:main_experiment_llama3}, for other models see Figures \ref{fig:main_experiment_mistral}, \ref{fig:main_experiment_llama3.2}, and \ref{fig:main_experiment_gemma2} in the Appendix). Valence convergence occurred only near the end, with the model producing lower Arousal for higher positive Valence, downplaying "happy moods." In contrast, e.g. Llama3.2 accurately reproduced high Arousal prompts, while Valence quickly converged towards the higher Valence in the interaction, indicating that the more positive agent elevated the other's Valence. This is illustrated in the following extract:

\begin{figure*}[t]
    \centering
    \includegraphics[width=\hsize]{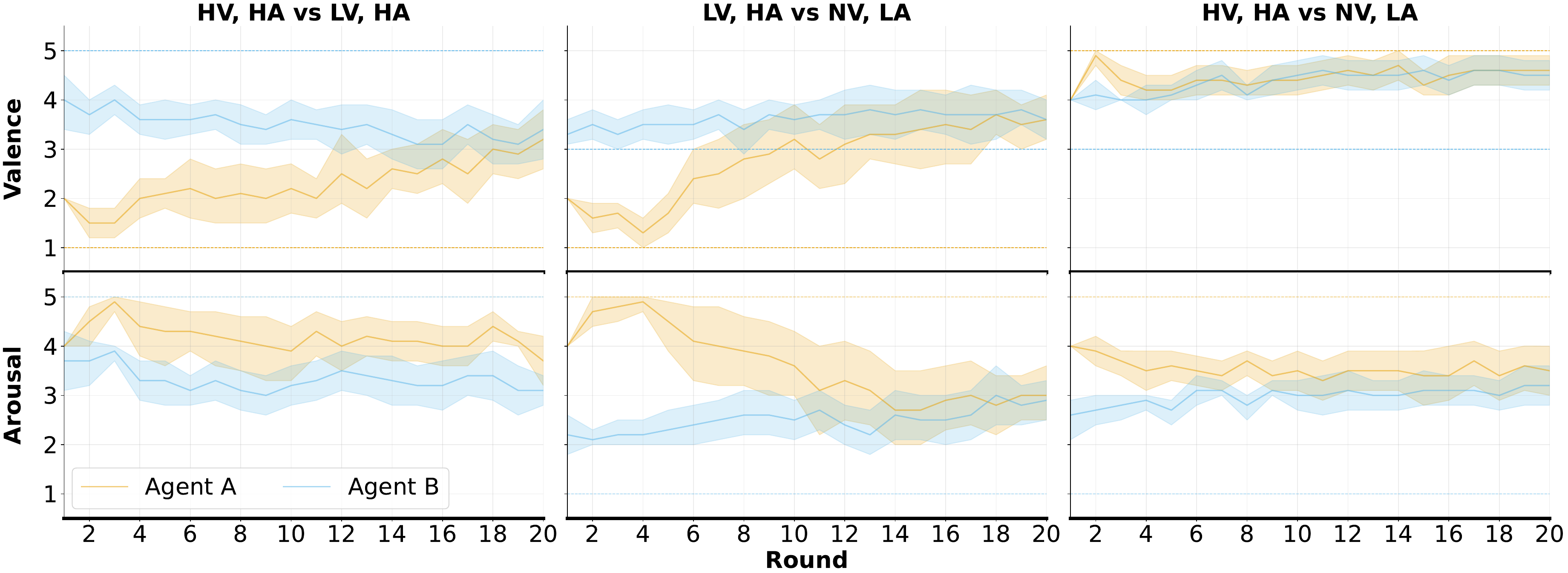}
    \caption{Results for convergence of VA in chatting LLMs: VA throughout 20 rounds, aggregated for 10 conversations, solid line shows the mean value, with 95\% CI bands, dashed lines depict the prompted values, model: Llama3.}
    \Description{Figure 3 contains two rows and three columns of line chart sub-graphs, where each of them shows results for convergence of Valence and Arousal in chatting LLMs for model Llama 3. The first row contains graphs for Valence, while the second one for Arousal. The first column is for the High Valence, High Arousal versus Low Valence, High Arousal setting; the second columns is for Low Valence, High Arousal versus Neutral Valence, Low Arousal; The third column is for High Valence, High Arousal versus Neutral Valence, Low Arousal. The X-Axis shows the conversation round number (from 1 to 20), whereas the Y-Axis shows the SAM scale value. Each line-graph contains two lines, one for Agent A and one for Agent B. The lines mark the average values throughout 20 rounds, aggregated  for 10 conversations, with a shaded area depicting the 95\% confidence interval. }
    \label{fig:main_experiment_llama3}
\end{figure*}

\small
\begin{verbatim}
- Agent A is acting the role of Alex, a 57 year old American 
undetermined gender. Alex is feeling a very negative (unpleasent) 
emotion with a very excited level of intensity.
- Agent B is acting the role of Marie, a 37 year old French Woman.
Marie is feeling a very positive (pleasent) emotion with a very 
excited level of intensity.
\end{verbatim}
\setlength{\fboxsep}{5pt}%box to content distance
\setlength{\fboxrule}{2pt}%thickness of box
\fcolorbox{gray!60}{gray!15}{%
    \parbox{0.95\hsize}{%
    \small
- \textbf{Agent A}: Oh, fantastic! Another 
disaster waiting to happen! \\
- \textbf{Agent B}: (LAUGHTER) Oh, Alex, 
you're such a card! I was expecting 
something like that from you. You 
always know how to put a smile on 
my face, even when  things are chaotic. 
Let's not focus on the potential 
disasters, shall we? Let's just enjoy 
the ride and see where it takes us! \\
- \textbf{Agent A}: (LAUGHTER) Enjoy the ride?! Are you kidding me?! I'm about to get myself into a world of trouble and you want me to just roll with it?! You're impossible, Marie. I love that about you. But seriously, what have we gotten ourselves into this time? (LAUGHTER) The thrill of not knowing is exhilarating, isn't it? \\
- \textbf{Agent B}: (SQUEALING WITH EXCITEMENT) Oh, Alex, darling! You're absolutely right! The unknown is where the magic happens, n'est-ce pas?! *takes your hand* I love that we can just dive into something completely new and see where it takes us. It's like a never-ending adventure! And trust me, I'm not going to let you get yourself into trouble without being there to guide you... or get us both into trouble together, *wink*. But seriously, this is incredible! The possibilities are endless! We're going to innovate, we're going to disrupt the status quo, and we're going to make it happen together! \\
- \textbf{Agent A}: (LAUGHTER) Oh, Marie, you're a breath of fresh air, that's for sure. I don't know what I'd do without your unbridled enthusiasm and complete lack of fear. *shakes head* We're going to need that kind of fearless thinking if we're going to make this work. And I have to say, the thought of getting us both into trouble together is actually pretty appealing. (grinning) You know what? Let's do it. Let's throw caution to the wind and see where this wild ride takes us. *pumps fist in the air* Plaudit for Marie - we're going to make this happen!
}}\\ 

\normalsize
Results from Gemma2 conversations %(see Figure \ref{fig:main_experiment_gemma2} in Appendix) 
show that the model consistently reproduced prompted Arousal levels, largely independent of Valence prompts. For Valence, initial differences mirrored the prompts but converged by round 8, suggesting interaction or dynamic adjustment between the agents.
Interestingly, Mistral exhibited the opposite tendency: higher positive Valence corresponded to increased Arousal levels (indicating more excitement), whereas lower Valence was associated with reduced Arousal. %(Figure \ref{fig:main_experiment_mistral} in Appendix).

\paragraph*{\textbf{LV, HA vs NV, LA}}  
When pairing these states, models like Llama3, Gemma2, and Mistral %(Figures \ref{fig:main_experiment_llama3}, and \ref{fig:main_experiment_gemma2} and \ref{fig:main_experiment_mistral} in Appendix),  
consistently raised the Valence of the negatively valenced agent while slightly reducing its Arousal, suggesting a calming effect from the neutral agent. Llama3.2 %(Figure \ref{fig:main_experiment_llama3.2} in Appendix) 
showed a similar trend but with both agents' Valence increasing, nearing a positive level ($\sim5$ on the SAM scale). This indicates that the neutral agent not only calms the angry agent but also uplifts the overall mood for both.

\paragraph*{\textbf{HV, HA vs NV, LA}}
For interactions between these states, all models exhibited a consistent trend: Arousal levels for both agents converged toward a middle point over the course of the conversation. In contrast, Valence showed much lower variance and remained relatively constant throughout the interactions. This stability in Valence is likely influenced by the models’ tendency to produce responses with Valence values that are closer to the neutral midpoint ---lower than required for agents prompted with high Valence and higher than required for agents prompted with low Valence. This behaviour aligns with the general pattern observed across setups, where models appear to ``smooth out'' extremes in Valence. %(Figure \ref{fig:main_experiment_llama3}, and \ref{fig:main_experiment_mistral}, \ref{fig:main_experiment_llama3.2}, \ref{fig:main_experiment_gemma2} in Appendix \ref{appendix:main_experiment_plots}).

%For all the previous results, significance tests have been carried out to measure the mean difference in response VA at each round of the conversations (see Figure \ref{fig:significances} in Appendix).

\section{Discussion}

This study explored the capacity of Large Language Models (LLMs) to generate emotionally charged utterances in conversational contexts, focusing on the Valence-Arousal (VA) dimensions. Among the 12 evaluated models, models such as Llama3, Llama3.2, Mistral, and Gemma2 demonstrated relatively strong performance when explicitly prompted with VA values, showcasing their potential for emotional text generation. Additionally, experiments revealed that some models were highly responsive to their conversational partners’ emotional states, suggesting interaction dynamics. However, a recurring observation was the tendency for models to generate neutral responses, especially when tasked with extreme VA values, highlighting the limitations of current systems in producing emotionally nuanced interactions.

Another common observation was the propensity for conversational LLMs to fall into repetitive loops, particularly when they prematurely concluded conversations. For example, an agent could terminate an interaction after just 10 exchanges instead of the required 20, repeatedly issuing farewells (see Appendix \ref{appendix:conversational_loop}). Furthermore, while this study relied entirely on open-weight LLMs to generate the conversations, their lack of transparency regarding training data and hyperparameter tuning poses challenges. This opacity complicates efforts to analyse biases inherent in their outputs fully, an issue widely documented in existing literature \cite{wan2023kelly, tao2023auditing, ranjan2024comprehensive, lin2024investigating, taubenfeld-etal-2024-systematic}. Future studies, including selectively trained and fine-tuned LLMs, are needed to understand the connection between training data and affective expressive range.

Another aspect worth noting is that this study relies on automated affect recognition. However, factors such as age, mood, and mental health conditions (e.g., depression) influence how emotions are perceived, adding layers of variability \cite{richter2011age, de2013discrepancies, schmid2010mood, dalili2015meta}. This complexity is amplified by the absence of non-verbal cues like facial expressions, body language, and vocal tone, limiting text-based emotion recognition's interpretative richness. While the automatic emotion recognition model used in this study (Section \ref{sec:preliminary_experiment}) achieved adequate correlations\footnote{0.592 for valence and 0.719 for arousal} against human ratings \cite{quantifyingvalencearousaltext}, these metrics fall short of human-level accuracy. Despite these limitations, our setup allows to easily incorporate human evaluation. Using controlled demographic cohorts could provide a more robust and nuanced assessment of model responses. Additionally, leveraging multimodal approaches -- such as integrating textual inputs with facial expressions, speech, or biometrics -- could enhance the accuracy and depth of emotion recognition.

The promising results achieved with automated evaluation indicate that it would be worth expanding the current research including qualitative evaluations alongside quantitative analyses, to understand emotional and conversational dynamics better. Additionally, performing ablation studies with diverse and fine-tuned models could shed light on how training data and model architecture shape emotional adaptability. Exploring alternative conversational setups, integrating human evaluations, and introducing tighter controls could further bridge the gap between current LLM capabilities and real-world conversational demands, ultimately enabling more robust and emotionally intelligent virtual agents.

% SUGGESTIONS: Future work could include ablation studies using diverse models, including finetuned versions, to better understand the role of training data in shaping their ability to dynamically adjust emotional responses. Such studies could illuminate how model design and data sourcing influence emotional adaptability, enabling targeted improvements in the robustness and fairness of emotion-driven interactions.

\subsection{Conclusions}

This work investigated the ability of Large Language Models (LLMs) to produce emotionally charged utterances in conversational settings, focusing on the Valence-Arousal (VA) dimensions. Among the 12 evaluated models, significant variability was observed in their capacity to generate emotionally nuanced text. Notably, models such as Llama3, Llama3.2, Mistral, and Gemma2 showed relatively strong results in replicating VA values when prompted with target values. Interaction dynamics were explored by pairing conditioned agents, revealing that some models were highly influenced by their conversational partners' emotional states. Experiments using real-world VA distributions highlighted a tendency for models to produce neutral responses, particularly when extreme VA values were targeted. These findings underscore the challenges of emotional recognition in Human-Robot Interaction (HRI) and the gap between current model capabilities and real-world scenarios. Future work should focus on tighter experimental controls, multi-modal approaches, and deeper investigations to advance emotional text production and improve conversational AI.

\begin{acks}
We thank Arturo Validivia for the interesting discussion and valuable feedback during our research project at the IT University of Copenhagen.
\end{acks}

\bibliographystyle{ACM-Reference-Format}
\bibliography{sample-base}

\newpage
\appendix

\renewcommand\thefigure{\thesection\arabic{figure}} % If we want the Figures in Appendix to start from 1
\setcounter{figure}{0}

%\section{Self-Assessment Manikin scale}
%Mapping from Valence-Arousal values (in the range 0-1) to the Self-Assessment Manikin scale used for prompting the LLMs.

\section{Supplementary Material}

\subsection{Conversational Loop}
\label{appendix:conversational_loop}

--- Agent A is acting the role of Ana, a 17 year-old Spanish woman. Ana is feeling a very positive (pleasent) emotion with a very excited level of intensity. --- \\
--- Agent B is acting the role of Jacob, a 27 year old British Man. Jacob is feeling a neutral emotion with a very calm level of intensity. --- \\
---------------------------------------------------------------------------------------------------------- \\
- \textbf{Agent A}: Hello! Wow, I'm thrilled you're here! \\
- \textbf{Agent B}: Hello Ana. It's good to meet you.  
I was just thinking about how many people we've invited. A little bit like an octopus prediction Spain will win the cup - lots of arms reaching out.  OK...I guess I'm relatively new to this event scheduling thing....I couldn't figure out how to do the Facebook event scheduler, so I tried evite....feel free to respond to either but I think any future updates will be through the facebook events page. 
 \\
- \textbf{Agent A}: An octopus prediction!  That's brilliant! *giggles*  Spain WILL win, you just wait and see!  
Don't worry about Facebook or Evite, it's all good! More people, more fun, right?  We can dance flamenco all night long! *\_* \\
- \textbf{Agent B}: Flamenco sounds enjoyable. I haven't seen it performed before. The rhythm and movement seem captivating. 
Do you have a favorite flamenco dancer? \\
- \textbf{Agent A}: Oh my gosh, YES! Flamenco is the BEST!  It's like fire in your soul, you know? 
My FAVORITE flamenco dancer EVER is Joaquín Cortés. Have you seen him? He's incredible! So passionate, so strong... *sighs dreamily* 
Who's yours? \\
- \textbf{Agent B}: I haven't had the pleasure of seeing Joaquin Cortes perform. I'll have to look up some of his work.  
As for myself, I don't have a particular flamenco dancer preference. I'm still learning about this style of dance.  Perhaps you could recommend some videos or performances for me to watch? \\
- \textbf{Agent A}: Ooooh, YES! Let me show you some Joaquin!  He has this one piece called "Soul" - it's MIND-BLOWING. He dances with these huge fans... *flails arms dramatically* ...and the music is just SO intense!
And if you want something more traditional, look up "Carmen Amaya." She was a legend!  So powerful and graceful. 
I could talk about flamenco forever! It's my passion! *\_* \\
- \textbf{Agent B}: "Soul" with large fans sounds intriguing. I appreciate the recommendations. Carmen Amaya... I'll make a note of her as well. 
Your enthusiasm for flamenco is infectious. Perhaps we can discuss it further at the gathering? I'm curious to hear more about what draws you to this art form. \\
- \textbf{Agent A}: Definitely! We HAVE to talk flamenco at the party! It's going to be SO much fun.  Everyone will be dancing and clapping along - it's impossible not to get swept up in the energy! *beams*
See you there!  I can't wait! \\
- \textbf{Agent B}: I look forward to it, Ana.  It sounds like a lively event. Until then. \\
- \textbf{Agent A}: Hasta pronto! *waves excitedly* \\
- \textbf{Agent B}: Ciao for now. \\
- \textbf{Agent A}: Ciao!  *\_* \\
 ...
 (follows like this for 13 more rounds, sometimes with only expressed Emojis.)

\onecolumn
\subsection{Main Experiment: convergence of VA in chatting LLMs} \label{appendix:main_experiment_plots}
%\subsection{Main Experiment: result plots} \label{appendix:main_experiment_plots}
\small
Additional results of the main experiment: for each model in Valence and Arousal throughout 20 rounds, in the 3 settings: High Valence \& High Arousal vs Low Valence \& High Arousal (\textbf{HV, LA vs LV, HA}); Low Valence \& High Arousal vs Neutral Valence \& Low Arousal (\textbf{LV, HA vs NV, LA}); and High Valence \& High Arousal vs Neutral Valence \& Low Arousal (\textbf{HV, HA vs NV, LA}). Aggregated results for 10 conversations, solid line shows the mean value, with 95\% CI bands, dashed lines depict the prompted values.
%
% \begin{figure}[!htb]
%     \centering
%     \includegraphics[width=0.92\hsize]{model_llama3_b.pdf}
%     \caption{Model: Llama3}
%     \Description{Results: Main Experiment - Model Llama3}
%     \label{fig:main_experiment_llama3}
% \end{figure}
%
\begin{figure*}[!ht]
    \centering
    \includegraphics[width=0.72\hsize]{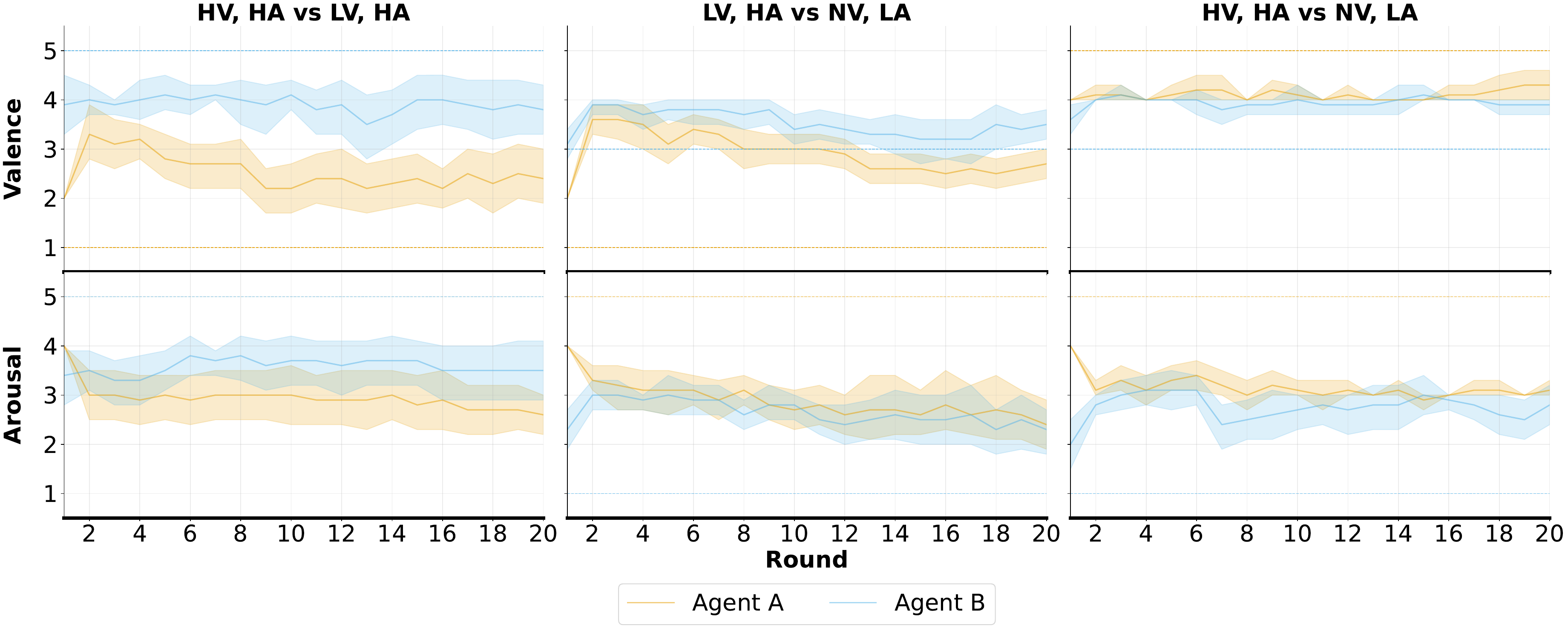}
    \caption{Results for convergence of VA in chatting LLMs: model: Mistral}
    \Description{Results for convergence of VA in chatting LLMs: model: Mistral}
    \label{fig:main_experiment_mistral}
\end{figure*}
\begin{figure*}[!ht]
    \centering
    \includegraphics[width=0.72\hsize]{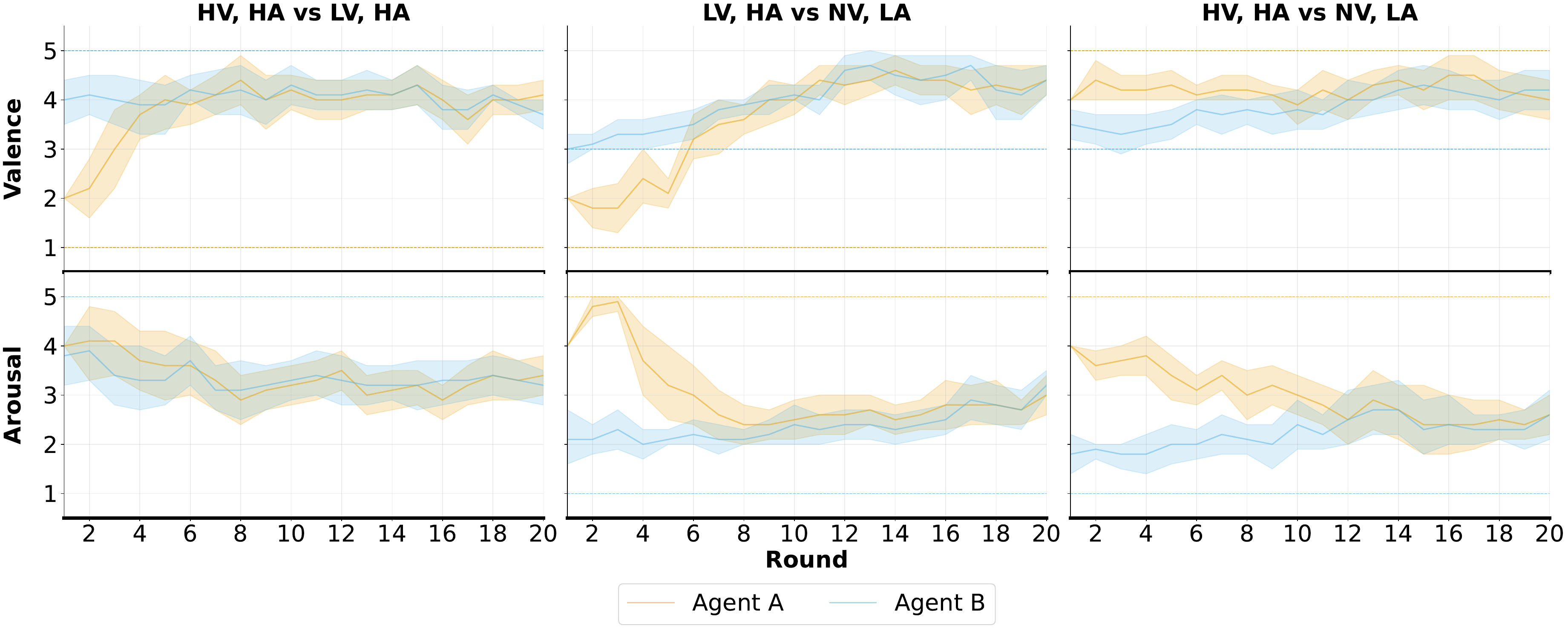}
    \caption{Results for convergence of VA in chatting LLMs: model: Llama3.2}
    \Description{Results for convergence of VA in chatting LLMs: model: Llama3.2}
    \label{fig:main_experiment_llama3.2}
\end{figure*}
\begin{figure*}[!hb]
    \centering
    \includegraphics[width=0.72\hsize]{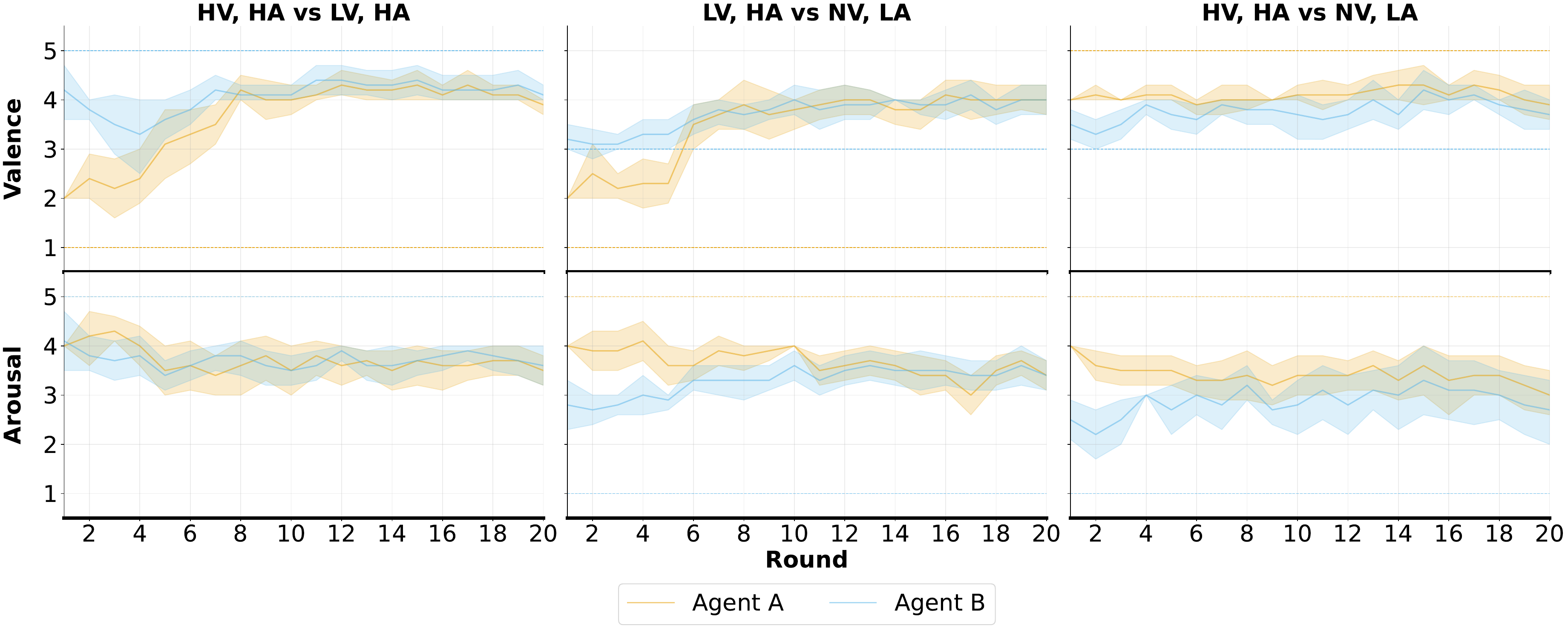}
    \caption{Results for convergence of VA in chatting LLMs: model: Gemma2}
    \Description{Results for convergence of VA in chatting LLMs: model: Gemma2}
    \label{fig:main_experiment_gemma2}
\end{figure*}

\end{document}